\documentclass[runningheads]{llncs}
\usepackage{graphicx}

\usepackage{acronym}
\usepackage{graphicx}
\usepackage{mathtools, amscd, mathrsfs}
\usepackage{multirow}
\usepackage{url}
\usepackage{tikz}
\usepackage{tikz-cd}

\usetikzlibrary{shapes.geometric}
\usepackage{amsmath}
\usepackage{amsfonts}
\usepackage{amsthm}
\usepackage{multirow}
\usepackage{colortbl}
\usepackage{bm}
\usepackage{array}

\allowdisplaybreaks

\def\eqref#1{equation~\ref{#1}}

\def\1{\bm{1}}

\def\vtheta{{\bm{\theta}}}

\def\vx{{\bm{x}}}

\DeclareMathAlphabet{\mathsfit}{\encodingdefault}{\sfdefault}{m}{sl}
\SetMathAlphabet{\mathsfit}{bold}{\encodingdefault}{\sfdefault}{bx}{n}

\def\gF{{\mathcal{F}}}

\def\gS{{\mathcal{S}}}

\def\gU{{\mathcal{U}}}

\def\gX{{\mathcal{X}}}
\def\gY{{\mathcal{Y}}}
\def\gZ{{\mathcal{Z}}}

\def\sR{{\mathbb{R}}}

\newcommand{\E}{\mathbb{E}}

\DeclareMathOperator*{\argmin}{arg\,min}

\DeclareMathOperator{\sign}{sign}

\newcommand{\ie}{i.e.}
\newcommand{\eg}{e.g.}

\newcommand{\eqq}{Eq.}

\usepackage{subcaption}
\usepackage{acronym}

\acrodef{ML}[ML]{Machine learning}
\acrodef{DL}[DL]{Deep learning}
\acrodef{DNN}[DNN]{Deep neural network}
\acrodef{NN}[NN]{Neural Network}
\acrodef{SVM}[SVM]{Support Vector Machine}
\acrodef{RL}[RL]{reinforcement learning}
\acrodef{ERM}[ERM]{Empirical Risk Minimization}
\acrodef{SVM}[SVM]{Support Vector Machine}
\acrodef{RNN}[RNN]{recurrent neural network}
\acrodef{LSTM}[LSTM]{long short-term memory}
\acrodef{PGD}[PGD]{projected gradient descent}
\acrodef{FGSM}[FGSM]{fast gradient sign method}

\acrodef{wrt}[\emph{w.r.t.}]{with respect to}
\acrodef{st}[\emph{s.t.}]{such that}

\begin{document}
\title{Learning to Learn from Mistakes: Robust Optimization for Adversarial Noise}
\titlerunning{Learning to Learn from Mistakes}

\author{Alex Serban\inst{1}\and
Erik Poll\inst{1}\and
Joost Visser\inst{2}
}
%
%
\institute{Radboud University, The Netherlands \and
Leiden University, The Netherlands \\
\email{a.serban@cs.ru.nl}}
\maketitle              
\begin{abstract}
Sensitivity to adversarial noise hinders deployment of machine learning algorithms in security-critical applications.
Although many adversarial defenses have been proposed, robustness to adversarial noise remains an open problem. 
The most compelling defense, adversarial training, requires a substantial increase in processing time and it has been shown
to overfit on the training data.
In this paper, we aim to overcome these limitations by training robust models in low data regimes and transfer adversarial knowledge between different models.
We train a meta-optimizer which learns to robustly optimize a model using adversarial examples and is able to transfer the knowledge learned to new models, without the need to generate new adversarial examples.
Experimental results show the meta-optimizer is consistent across different architectures and data sets, suggesting it is possible to automatically patch adversarial vulnerabilities. 
\end{abstract}
%
%
%
\section{Introduction}
\label{sec:intro}

\ac{ML} algorithms exhibit low robustness against intentionally crafted perturbations~\cite{goodfellow2014explaining} or high invariance for distinct inputs~\cite{jacobsen2018excessive}.
From a security standpoint this means an attacker can craft inputs that look similar but cause \ac{ML} algorithms to misbehave or find distinct inputs that give the same result.
From a safety standpoint, this means that \ac{ML} algorithms are not robust against perturbations close to an input or are inflexible to changes in the operational environment.

Creating defenses (or finding robust counterparts for  \ac{ML} models) has received increased attention, in particular in the field of adversarial examples.
Although many defenses have been proposed, with the most notable results using formal methods to guarantee robustness~\cite{huang2018safety} most solutions overfit on the training data and behave poorly against data outside this distribution~\cite{zhang2019limitations,rice2020overfitting}.
Theoretical investigations suggest these results are expected because training robust models requires more data~\cite{cullina2018pac}, more computational resources~\cite{bubeck2018adversarial} or accepting a trade off between accuracy and robustness~\cite{tsipras2018there}.
Moreover, solutions to one vulnerability have a negative impact on others~\cite{jacobsen2019exploitint}.

On a different path, designing \ac{ML} algorithms capable to rapidly adapt to changes in the operational environment, to adapt to distribution shifts or capable to learn from few samples is an active research field.
Particularly, the field of meta-learning investigate optimization algorithms learned from scratch for faster training~\cite{andrychowicz2016learning}, with less resources~\cite{ravi2016optimization} and capable of rapid adaptation~\cite{finn2017model}.

In this paper, we show that meta-learning algorithms can be used to extract knowledge from a model's vulnerability to adversarial examples and transfer it to new models.
We train a meta-optimizer to learn how to robustly optimize other models using adversarial training. 
Later, when asked to optimize new models without seeing adversarial examples,  the trained meta-optimizer can do it robustly.
This process is analogous to learning a regularization term for adversarial examples, instead of manually designing one. 
The experimental results suggest a broader horizon, in which algorithms learn how to automatically repair or treat vulnerabilities without explicit human design.

The rest of the paper is organized as follows.
In Section~\ref{sec:background} we introduce prerequisites and related work.
Section~\ref{sec:formal} formalizes meta-learning and the adversarial training problems and gives implementation details.
Section~\ref{sec:experiments} presents experimental results on two distinct data sets, followed by a discussion in Section~\ref{sec:discussion} and conclusions in Section~\ref{sec:conclusions}.

\section{Background and Related Work}
\label{sec:background}
We focus on the task of supervised classification, \ie~given a set of inputs from the input space $\gX$ and a set of labels from the space $\gY$ 
 a \ac{ML} algorithm attempts to find a mapping from $f: \gX \rightarrow \gY$ which minimizes the number of misclassified inputs.
We assume that $\gX$ is a metric space such that a distance function $d(\cdot)$ between two points of the space exists.
The error made by a prediction $f(\vx) = \hat{y}$ when the true label is $y$ is measured by a loss function $l:\gY \times \gY \rightarrow \sR$ with non-negative values when the labels are different and zero otherwise.
$f(\cdot)$ is defined over a hypothesis space $\gF$ which encompasses any mapping from $\gX$ to $\gY$ and can take any form~--~\eg~a linear function or a neural network.
Through learning, a \ac{ML} algorithm selects $f^{*}(\cdot)$ from $\gF$ such that the expected empirical loss on a training set $\gS$ consisting of pairs of samples $(\vx_i, y_i) \sim \gZ = \gX \times \gY$ is minimal. 

A robust solution to the minimization problem above involves immunizing it against uncertainties in the input space.
In the adversarial examples setting, uncertainties are modeled in the space around an input $\vx_u$: $\gU_{\vx} = \{\vx | d(\vx, \vx_c) \leq \epsilon\}$, where $d(\cdot)$ is a distance function defined on the metric space $\gX$.
The robust counterpart of the learning problem is defined as:
\begin{equation}
f = \argmin_{f \in \gF} \E_{(\vx, y) \sim \gS} [ \max_{\vx_u \in \gU_{\vx}} l(f(\vx_u), y)],
\label{eq:rerm}
\end{equation}
\noindent
where $\vx_u$ is a realization of $\vx$ in the uncertainty set described by $d(\cdot)$.
A common distance function used in the field of adversarial examples is the $l_p$-norm.
Arguably, this metric can not capture task-specific semantic information, but it is suitable for comparative benchmarks.
Moreover, the lack of robust p-norm solutions for complex tasks makes it hard to believe that other notions of robustness can lead to better results~\cite{oneval}.
In this paper we use the $l_{\infty}$-norm distance. 


Finding models robust to adversarial examples is an open question, spanning two  research directions: 
(1) finding solutions robust to an upper bound on the perturbation size $\epsilon$ (\ie~no perturbation higher than $\epsilon$ can cause a misclassification)~\cite{goodfellow2014explaining,madry2017towards}  
and (2) finding lower bounds on robustness (\ie~no adversarial example can be found in a space around an input)~\cite{wong2017provable,mirman2018differentiable}.

Notable results are obtained by solving the inner maximization problem from \eqq~(\ref{eq:rerm}) with \ac{PGD}~\cite{madry2017towards}, by training with over-approximations of the input~\cite{mirman2018differentiable} or extensively searching for spaces where no adversarial examples can be found using exact solvers~\cite{katz2017reluplex}.
In all cases, the benchmarks show overfitting on the training distribution and can be bypassed by samples outside it~\cite{zhang2019limitations}. 

Until recently, solving the outer minimization objective from \eqq~(\ref{eq:rerm}) relied on static, hand designed algorithms, such as stochastic gradient descent or ADAM. 
This line of research is driven by the no free lunch theorem of optimization 
which states that, on average, in combinatorial optimization no algorithm can do better than a random strategy; suggesting that designing specific algorithms for a class of problems is the only way to improve performance.

Recent advancements in the field of meta-learning have taken a different approach, posing the problem of algorithm design dynamically and modeling it with neural networks~\cite{andrychowicz2016learning} or as a \ac{RL} problem~\cite{li2016learning}.
In both cases, the algorithms show empirically faster convergence and the ability to adapt to different tasks.

For example, in~\cite{andrychowicz2016learning}~the hand designed update rules are replaced with a parameterized function modeled with a \ac{RNN}.
During training, both the  \emph{optimizer} and the \emph{optimizee} parameters  are adjusted.
The algorithm design now becomes a learning problem, allowing to specify constraints through data examples.
%
The method has roots in previous research~\cite{schmidhuber1993neural}, where initial \acp{RNN} were designed to update their own weights. 
The results of backprobagation from one network were only later fed to a second network which learns to update the first~\cite{younger2001meta}.
Andrychowicz et al.~build on previous work using a different learning architecture.
Similarly,~\cite{ravi2016optimization} used this paradigm to train neural networks in a few shot regime and~\cite{santoro2016meta} augmented it with memory networks.
Meta-learning has also shown promising results in designing algorithms for fast adaptation using gradient information~\cite{finn2017model}.

Instead of using \acp{RNN}, \cite{li2016learning}~formulate the optimization problem as a \ac{RL} problem where the reward signal represents successful minimization and train an agent using guided policy search.
Later, the authors refine their method for training neural networks~\cite{li2017learning}.
In both cases the agent learns to optimize faster than hand designed algorithms and exhibits better stability.

Recent research in adversarial examples has also tackled the need to decrease training resources by either accumulating
perturbations~\cite{shafahi2019adversarial,wong2020fast} or by restricting back-propagation to some layers of the model~\cite{zhang2019you}.
While the latter method requires forward and backward passes, the former  reduces the need to do backward passes in order to generate adversarial examples. 
and not to abstract or transfer information from adversarial training.





\section{Learning to Optimize Robustly}
\label{sec:formal}

Meta-learning frames the learning problem at two levels: acquiring higher level (meta) information about the optimization landscape and applying it to optimize one task.
In this paper, we are interested to learn robust update rules and transfer this knowledge to new tasks without additional constraints.
We focus on training robust \ac{ML} models through adversarial training, which is one of the most effective defenses against upper bounded adversarial examples~\cite{madry2017towards}.
Because generating adversarial examples during training is time consuming, especially for iterative procedures and can only provide robustness for the inputs used during training~\cite{zhang2019limitations}, through meta-learning we learn to optimize robustly without explicit regularization terms and transfer the knowledge to new tasks, without the need to generate new adversarial examples. 

At a high level, adversarial training discourages \emph{local} sensitive behavior~--~in the vicinity of each input in the training set~--~by guiding the model to behave constantly in regions surrounding the training data.
The regions are defined by a chosen uncertainty set (as in \eqq~(\ref{eq:rerm})).
This procedure is equivalent to adding a prior that defines local constancy, for each model we want to train.
In most cases, specifying this prior is not trivial and requires the design of new regularization methods~\cite{miyato2015distributional}, new loss functions~\cite{wong2017provable} or new training procedures~\cite{mirman2018differentiable}.
In this paper we take a different approach and try to learn a regularization term automatically, using meta-learning.
During the meta-knowledge acquisition phase, the meta-optimizer learns to perform the updates robustly using adversarial training.
Later, the knowledge acquired is transferred to new models using the meta-optimizer to train new models, without generating adversarial examples.
In the next paragraphs we describe the meta-optimizer, some implementation details and the adversarial training procedure.

\paragraph{Learning to Optimize.}
The classification function defined in Section~\ref{sec:background}, $f(\cdot)$, is parametrized by a set of parameters $\theta$.
Upon seeing new data, we update the parameters in order to minimize the prediction errors.
The update  consists in moving one step in the opposite direction of the gradient:
\begin{equation}
\theta_{t+1} = \theta_{t} -  \eta \nabla_{\theta_{t}} l(\cdot),
\label{eq:update}
\end{equation}
where $\eta$ (the learning rate) determines the size of the step.
Different choices of $\eta$ or ways to automatically adapt it results in different optimization algorithms such as stochastic gradient descent or ADAM.

In order to avoid overfitting or impose additional constraints such as constancy around inputs, it is common to add a regularization term to the loss function which will be back-propagated and reflected on all parameter updates.
Instead of looking for regularization terms manually, we use a method to automatically learn robust update steps with regularization included. 

As discussed in Section~\ref{sec:background}, a parameterized update rule has been previously represented with a \ac{RNN}~\cite{andrychowicz2016learning,ravi2016optimization} or as a \ac{RL} problem~\cite{li2016learning}.
In this paper, we follow an approach similar to~\cite{andrychowicz2016learning} and model the update rule with a \ac{RNN} with \ac{LSTM} cells:

\begin{equation*}
\theta_{t} = \theta_{t-1} + c_{t},
\end{equation*}
\noindent
where :
\begin{equation}
c_{t} =  f_{t}  c_{t-1} + i_{t-1}  \tilde{c}_{t},
\label{eq:update_lstm}
\end{equation}
is the output of an \ac{LSTM} network $m$ with input $\nabla_{\theta_{t}}(l(\cdot))$:
\begin{gather}
\begin{bmatrix} c_t \\ h_{t+1} \end{bmatrix}
=
m(\nabla_t, h_t, \phi),
\label{eq:lstm_internal}
\end{gather}
and $\phi$ are the LSTM's parameters.

In \cite{ravi2016optimization}, the authors consider each term in~\eqq~(\ref{eq:update_lstm}) equivalent to each term in~\eqq~(\ref{eq:update})~--~\eg~$f_{t} =1$, $c_{t-1}=\theta_{t}$~--~and disentangle the internal state of the \ac{LSTM}, $h_t$, with special terms for individual updates of $f_t$ and $i_t$.
This type of inductive bias brings benefits in some cases and will be further discussed in Section~\ref{sec:discussion}.
However, in this paper we try to avoid such biases whenever possible.

\paragraph{Parameters Sharing and Gradient Preprocessing.}
In order to limit the number of parameters of the optimizer, we follow a procedure similar to~\cite{andrychowicz2016learning,ravi2016optimization} in which for each parameter of the function we want to optimize, $\theta_{i:n}$, we keep an equivalent internal state of the optimizer, $h_{i:n}$, but share the weights $\phi$ between all states.
This procedure allows a more compact optimizer to be used and makes the update rule dependent only on its respective past representation, thus being able to simulate hand designed optimization features such as momentum.

Moreover, since gradient coordinates can take very distinct values, we apply a common normalization step in which the gradients are scaled and the information about their magnitude and their direction is separated:
\begin{align}
\nabla &\rightarrow
\begin{cases}
\left(\frac{\log(|\nabla|)}{p}\,, \sign(\nabla)\right)& \text{if } |\nabla| \geq e^{-p}\\
(-1, e^p \nabla)             & \text{otherwise.}
\end{cases}
\label{eq:process}
\end{align}
We experiment with different values for $p$ by grid search and observe that increasing the size of $p$ yields better results when the perturbations are larger.
However, for consistency, we use $p=10$ for all experiments.

The meta-optimizer's parameters are updated using an equivalent to \eqq~(\ref{eq:update}).
Since its inputs are based on the gradient information of the function to be optimized (also called the optimizee) the updates will require second order information about it (taking the gradient of the gradient). 
This information is commonly used for meta-learning~--~\eg~in~\cite{finn2017model,wichrowska2017learned}~--~and will be further discussed in Section~\ref{sec:discussion}.
However, in this paper only first order information is used, corresponding to limiting the propagation of the  optimizer's gradient on the optimizee parameters (or stopping the gradient flow in the computational graph).

\paragraph{Adversarial Training.}
Adversarial training is still one of the most effective defenses against adversarial examples.
This procedure is equivalent to adding a regularization term to the loss function of the optimizee, corresponding to:
\begin{equation}
\tilde{l}(\cdot) =    \alpha  l(\vtheta, \vx, y) +  (1-\alpha)  l(\vtheta, \vx', y),
\label{eq:adv_training}
\end{equation}
\noindent
where $\vx'$ is an adversarial example generated from input $\vx$ and $\alpha$ is a parameter which controls the contribution of the adversarial loss.
In some cases adversarial training is performed using only the adversarial loss, corresponding to $\alpha = 0 $ in \eqq~(\ref{eq:adv_training}), \eg~when training with the worst adversarial loss as in~\eqq~(\ref{eq:rerm}).

Several methods to generate adversarial examples have been proposed, ranging from fast, less precise, attacks~\cite{goodfellow2014explaining} to strong, adaptive attacks~\cite{carlini2017towards}.
However, due to processing constraints, only some algorithms are used in adversarial training: the \ac{FGSM} attack~--~which moves one step in the direction of the gradient~--~and the \ac{PGD} attack~--~which approximates the uncertainty set $\gU_{\vx}$ running several steps of the \ac{FGSM} and projecting the outcomes on the norm ball surrounding an input, defined by $\epsilon$.

%
In this paper, we use both \ac{FGSM} and \ac{PGD} during the meta-learning  phase to generate adversarial examples and incorporate them in the training procedure.
Formally, the \ac{FGSM} attack is defined as:
\begin{equation}
\vx' = \vx + \epsilon \sign(\nabla_x l(\vtheta, \vx, y)).
\label{eq:fgsm}
\end{equation}
The term $\epsilon$ controls the size of the perturbation added in order to cause a misclassification.
We are aware that defenses against \ac{FGSM} attacks sometimes lead to a false sense of protection and to gradient masking~\cite{athalye2018obfuscated}, however, we remind that hereby the goal is to reduce the data needed to train robust models and increase the generalization outside the training distribution.

The \ac{PGD} attack is defined as follows:
\begin{equation}
\vx' = \prod_{n} (\vx' + \epsilon \sign( \nabla_xl(\vtheta, \vx, y)) ) .
\label{eq:pgd}
\end{equation}
\noindent
If the number of iteration suffices to approximate the space we want to provide robustness to, training with adversarial examples generated with this method should protect against any perturbation in the space defined by $\epsilon$.
In order to obtain better approximations, we can increase the iteration number, $n$.
However, this comes at increased computational costs.


\section{Results}
\label{sec:experiments}

In all experiments the optimizer consists of a two-layer \ac{LSTM} network with a hidden state size of 20.
We compare the results on training two types of neural networks on two distinct data sets with the adaptive optimizer ADAM.

We focus on two experiments related to training neural networks, as in prior work on meta-learning~\cite{andrychowicz2016learning,ravi2016optimization,li2016learning,li2017learning}. 
More experiments with minimizing other functions~--~\eg~logistic regression~--~and an integration with the Cleverhans framework are available in the project's repository, which can be found at~https://github.com/NullConvergence/Learning2LearnAdv.
In all cases, an optimizer is trained using normal and adversarial examples on a data set and tested by training a robust optimizee without generating adversarial examples.
Several perturbation sizes are analyzed, as introduced below.

\subsection{MNIST}
\label{subsec:mnist}
\begin{figure*}[t!]
	\centering
	\begin{subfigure}[t]{.23\textwidth}
		\centering
		\includegraphics[width=3.1cm, keepaspectratio]{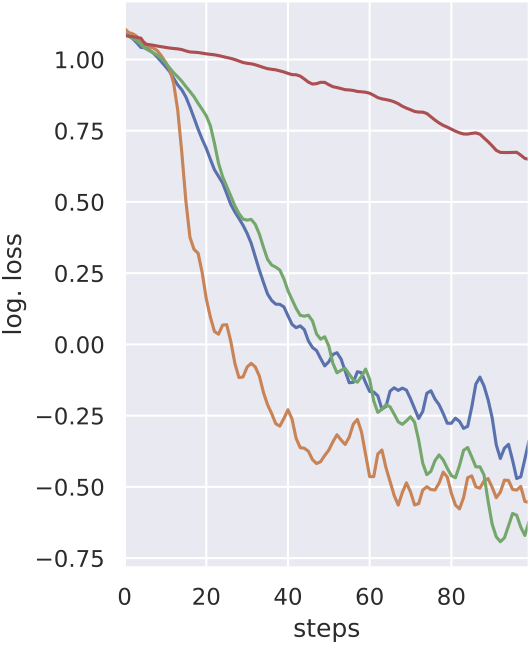}
		\caption{$\epsilon$ = 0.05}
		\label{fig:mnist_fgsm:a}
	\end{subfigure}
	\hfill\hfill\hfill\hfill\hfill\hfill\hfill\hfill\hfill
	\begin{subfigure}[t]{.23\textwidth}
		\centering
		\includegraphics[width=2.45cm, keepaspectratio]{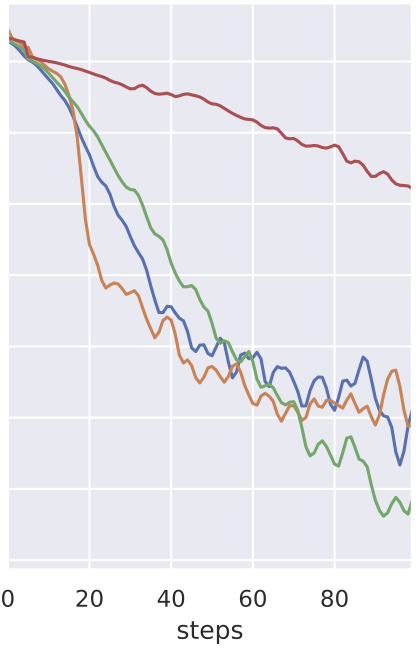}
		\caption{$\epsilon$ = 0.1}
		\label{fig:mnist_fgsm:b}
	\end{subfigure}
	\hfill
	\begin{subfigure}[t]{.23\textwidth}
		\centering
		\includegraphics[width=2.45cm, keepaspectratio]{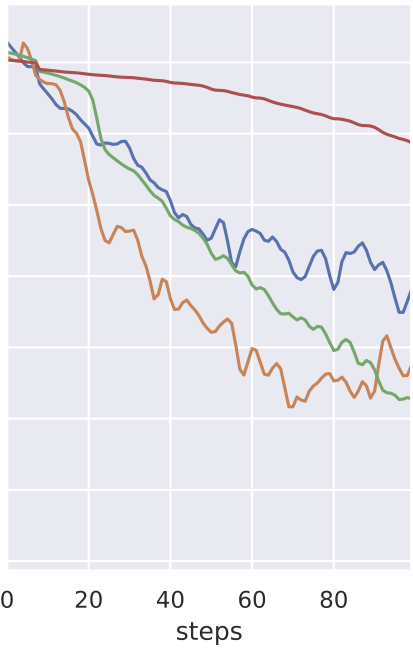}
		\caption{$\epsilon$ = 0.2}
		\label{fig:mnist_fgsm:c}
	\end{subfigure}
	\hfill
	\begin{subfigure}[t]{.23\textwidth}
		\centering
		\includegraphics[width=4cm, keepaspectratio]{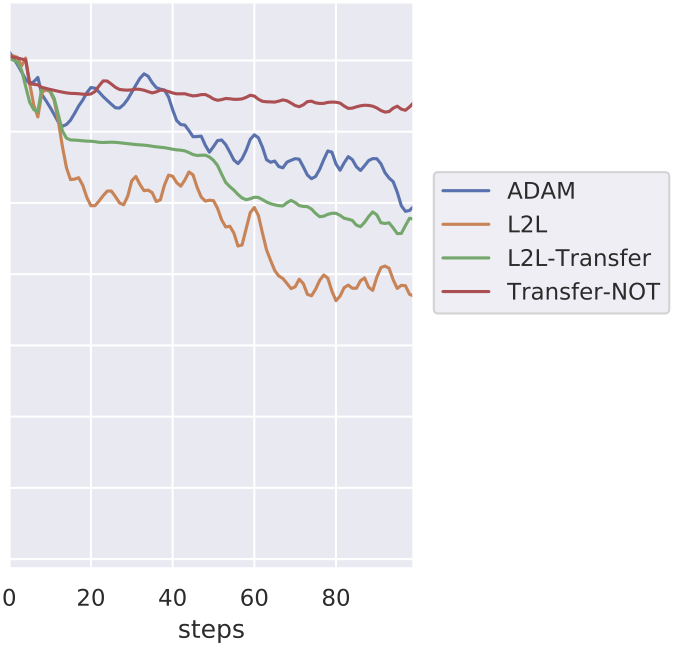}
		\caption{$\epsilon$ = 0.3}
		\label{fig:mnist_fgsm:d}
	\end{subfigure}
	\hfill
	\caption{The loss landscape when training a neural network on the MNIST data set, perturbed with FGSM and different perturbation sizes ($\epsilon$).
		The meta-optimizer is trained with adversarial examples~--~label L2L~--~transferred to a scenario where training is performed with normal and adversarial data, but tested with adversarial examples~--~label L2L-Transfer~--~and compared with a meta-optimizer trained with normal data and transferred to adversarial settings~--~label Transfer-NOT~--~and with ADAM. Best seen in color.}
	\label{fig:mnist_fgsm}
\end{figure*}

\begin{figure*}[t!]
	\centering
	\begin{subfigure}[t]{.23\textwidth}
		\centering
		\includegraphics[width=3.1cm, keepaspectratio]{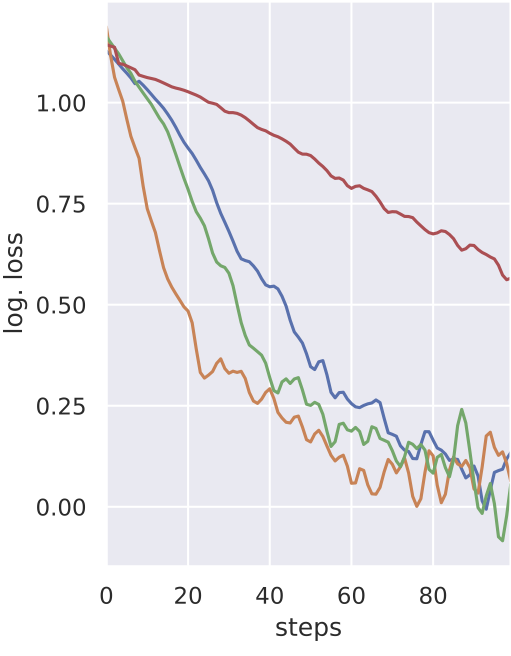}
		\caption{$\epsilon$ = 0.05}
		\label{fig:mnist_pgd:a}
	\end{subfigure}
	\hfill
	\begin{subfigure}[t]{.23\textwidth}
		\centering
		\includegraphics[width=2.45cm, keepaspectratio]{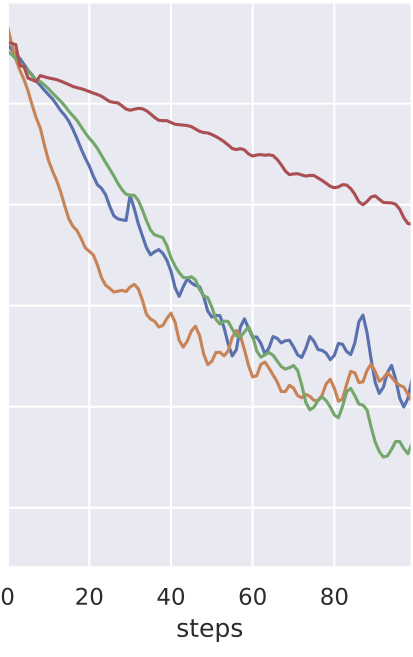}
		\caption{$\epsilon$ = 0.1}
		\label{fig:mnist_pgd:b}
	\end{subfigure}
	\begin{subfigure}[t]{.23\textwidth}
		\centering
		\includegraphics[width=2.45cm, keepaspectratio]{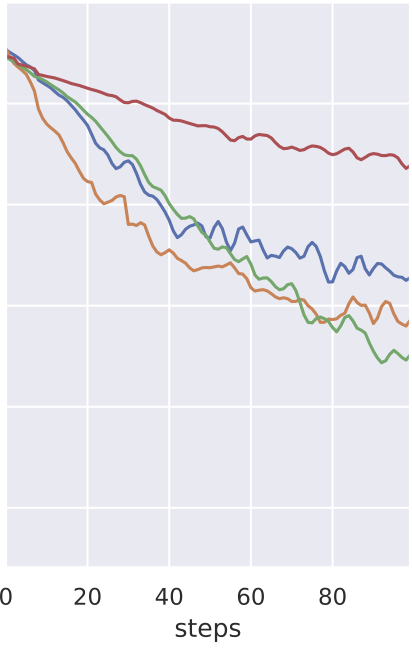}
		\caption{$\epsilon$ = 0.2}
		\label{fig:mnist_pgd:c}
	\end{subfigure}
	\hfill
	\begin{subfigure}[t]{.23\textwidth}
		\centering
		\includegraphics[width=4cm, keepaspectratio]{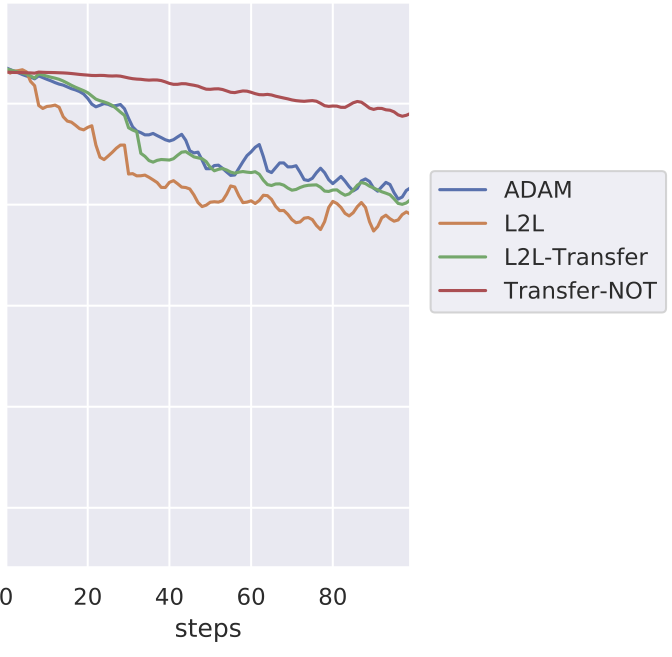}
		\caption{$\epsilon$ = 0.3}
		\label{fig:mnist_pgd:d}
	\end{subfigure}
	\hfill\hfill\hfill
	\caption{The loss landscape when training a neural network on the MNIST data set perturbed with the PGD method and different perturbation sizes ($\epsilon$). The legend is detailed in the caption of Figure~\ref{fig:mnist_fgsm}.}
	\label{fig:mnist_pgd}
\end{figure*}

We begin by training a small, fully connected, neural network with 20 units and ReLU activation on the MNIST data set.
The perturbations take different values in the set $\epsilon \in \{0.05,0.1,0.2,0.3\}$ for both attacks introduced earlier (\eqq~(\ref{eq:fgsm}) and \eqq~(\ref{eq:pgd})).
We experiment with different learning rates by grid search and find the best to be $0.001$  for the meta-optimizer.
Training is performed using the common cross entropy loss function, with a batch size of 128.
We shuffle the training data set (consisting of 60.000 examples) and divide in two parts equally: the first is used to train a meta-optimizer using both normal data and adversarial examples and the second is used to test its performance while training with normal data and testing with perturbed data.
Each experiment ran for 100 steps and the average results are illustrated in Figures~\ref{fig:mnist_fgsm}~and~\ref{fig:mnist_pgd}.
All experiments are done using $\alpha=0.5$ in \eqq~(\ref{eq:adv_training}) during training and $\alpha=0.0$ during the meta-optimizer transfer phase, as first introduced in~\cite{goodfellow2014explaining}.
In addition to ADAM's performance compared to the meta-optimizer, we evaluate the performance of the meta-optimizer during training and the performance of training a meta-optimizer using $\alpha=1$ and testing with $\alpha=0 $ (L2L and Transfer-NOT labels in Figures~\ref{fig:mnist_fgsm}~and~\ref{fig:mnist_pgd}).
Figure~\ref{fig:mnist_fgsm} illustrates the results from generating adversarial examples using the \ac{FGSM} method (\eqq~(\ref{eq:fgsm})). 
The loss functions start from approximately the same value because the networks are always initialized with the same values.

%

In all cases, the meta-optimizer is able to transfer the information learned during training and has comparable performance to ADAM (in some cases performing better).
We remind that during testing the optimizer uses normal data, but the plots are generated by feeding adversarial perturbed data to the optimizee.
This implies that the meta-optimizer proposes update rules which lead to smooth surfaces around the tested inputs. 
Moreover, it is able to learn a robust regularization term during training and transfer it to new tasks without the need to generate new data.
Also, the trained meta-optimizer exhibits more stable behavior.
This brings evidence that adversarial training leads to more interpretable gradients~\cite{tsipras2018there}.

When the optimizer is trained only with normal examples, but used to optimize the model using adversarial examples ~--~Transfer-NOT label in Figure~\ref{fig:mnist_fgsm}~--~its performance decrease significantly.
This implies that a meta-optimizer is domain specific and does not have the general behavior of ADAM, an observations which will be further discussed in Section~\ref{sec:discussion}.


In Figure~\ref{fig:mnist_pgd} we illustrate the results from running similar experiments, but generate adversarial examples using the \ac{PGD} method from \eqq~(\ref{eq:pgd}).
Training with \ac{PGD} is generally performed only using  the perturbed examples (corresponding to $\alpha=0$ in \eqq~(\ref{eq:adv_training})), as in the original paper~\cite{madry2017towards}.
We take a similar approach in this paper.

The results are consistent with the \ac{FGSM}  method, although the gap between ADAM and the transferred meta-optimizer is smaller. 
A constant decrease in performance is also observed, possibly corresponding to the decrease in performance specific to adversarial training~\cite{tsipras2018there}.
Nevertheless, the results are consistent and bring evidence that the meta-optimizer is able to learn robust update rules.

\subsection{CIFAR-10}
\label{sec:cifar}

\begin{figure*}[t!]
	\centering
	\begin{subfigure}{.23\textwidth}
		\centering
		\includegraphics[width=3.1cm, keepaspectratio]{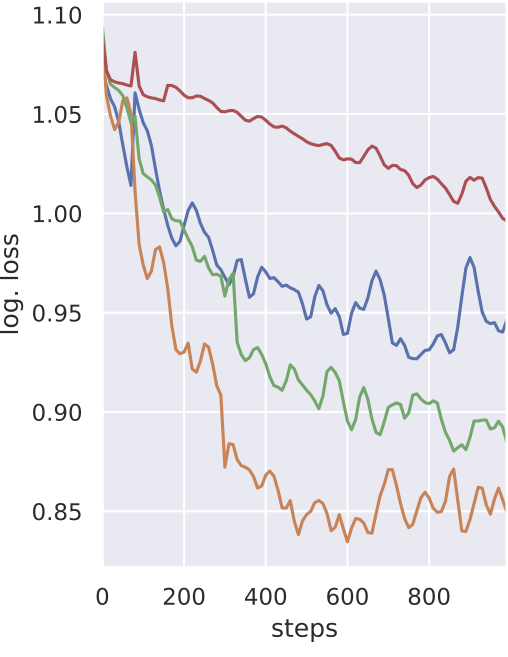}
		\caption{$\epsilon$ = 0.05}
		\label{fig:cifar_fgsm:a}
	\end{subfigure}
	\hfill
	\begin{subfigure}{.23\textwidth}
		\centering
		\includegraphics[width=2.45cm, keepaspectratio]{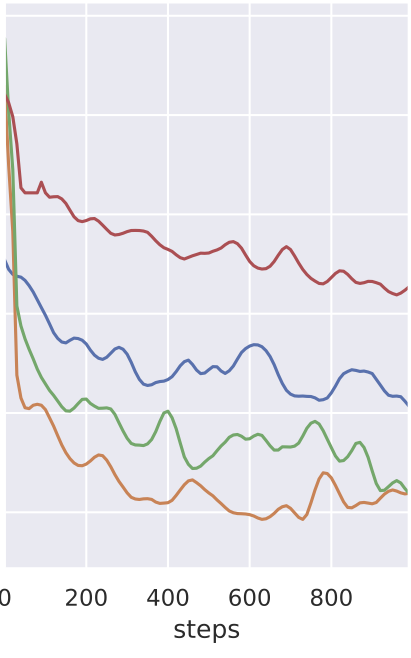}
		\caption{$\epsilon$ = 0.1}
		\label{fig:cifar_fgsm:b}
	\end{subfigure}
	\begin{subfigure}{.23\textwidth}
		\centering
		\includegraphics[width=2.45cm, keepaspectratio]{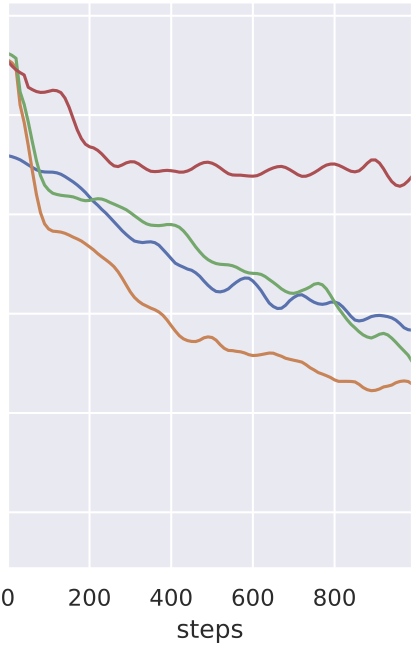}
		\caption{$\epsilon$ = 0.2}
		\label{fig:cifar_fgsm:c}
	\end{subfigure}
	\hfill
	\begin{subfigure}{.23\textwidth}
		\centering
		\includegraphics[width=4cm, keepaspectratio]{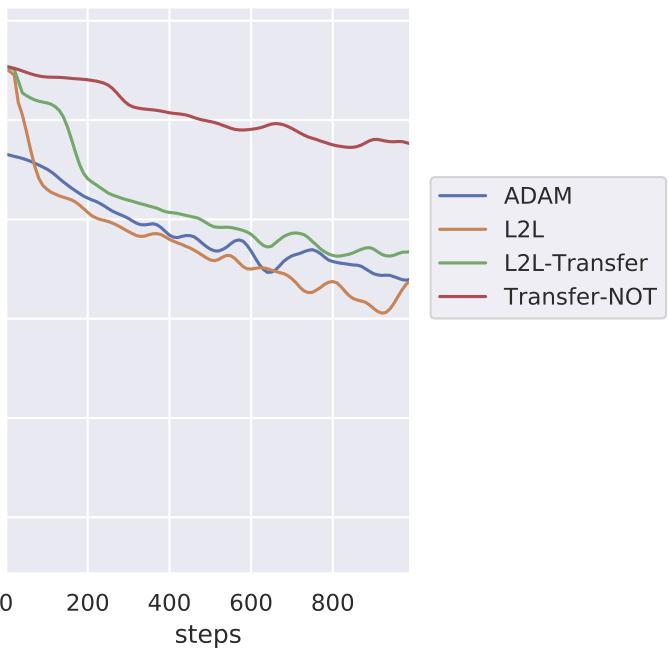}
		\caption{$\epsilon$ = 0.3}
		\label{fig:cifar_fgsm:d}
	\end{subfigure}
	\hfill\hfill\hfill
	\caption{The loss landscape when training a neural network on the CIFAR-10 data set, perturbed with the FGSM method and different perturbation sizes ($\epsilon$).
		The legend is detailed in the caption of Figure~\ref{fig:mnist_fgsm}.}
	\label{fig:cifar_fgsm}
\end{figure*}
%
\begin{figure*}[t!]
	\centering
	\begin{subfigure}{.23\textwidth}
		\centering
		\includegraphics[width=3.1cm, keepaspectratio]{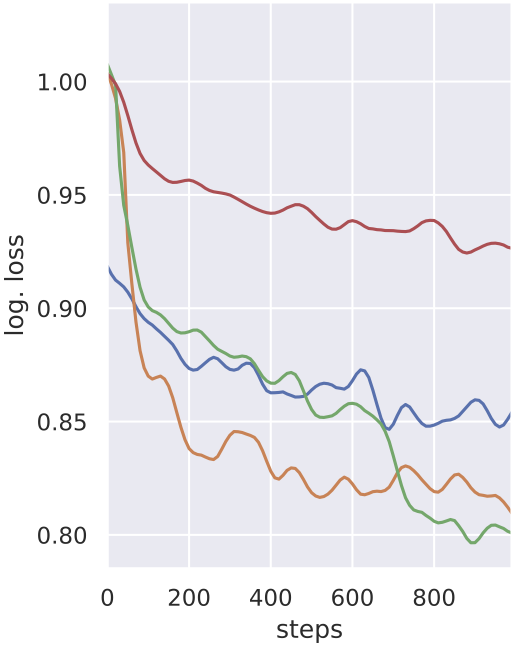}
		\caption{$\epsilon$ = 0.05}
		\label{fig:cifar_pgd:a}
	\end{subfigure}
	\hfill
	\begin{subfigure}{.23\textwidth}
		\centering
		\includegraphics[width=2.45cm, keepaspectratio]{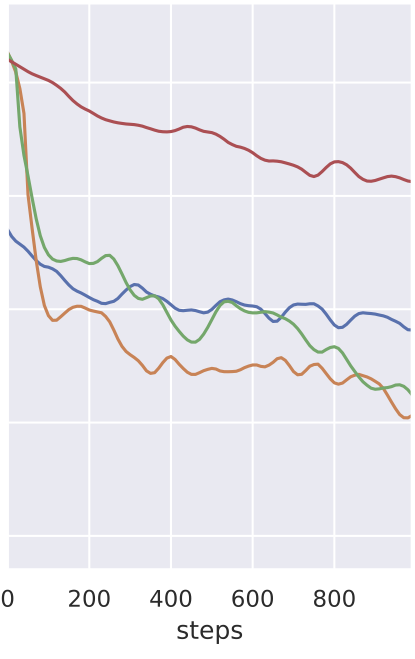}
		\caption{$\epsilon$ = 0.1}
		\label{fig:cifar_pgd:b}
	\end{subfigure}
	\begin{subfigure}{.23\textwidth}
		\centering
		\includegraphics[width=2.45cm, keepaspectratio]{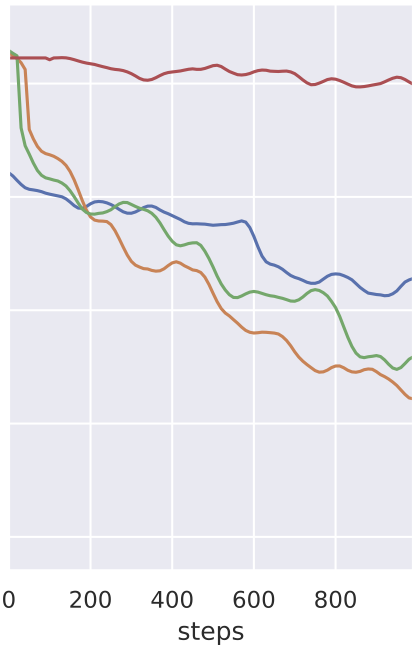}
		\caption{$\epsilon$ = 0.2}
		\label{fig:cifar_pgd:c}
	\end{subfigure}
	\hfill
	\begin{subfigure}{.23\textwidth}
		\centering
		\includegraphics[width=4cm, keepaspectratio]{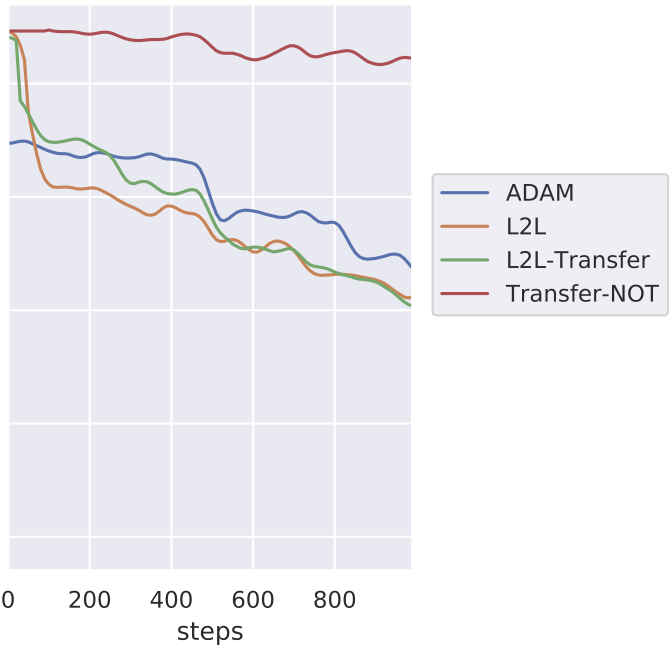}
		\caption{$\epsilon$ = 0.3}
		\label{fig:cifar_pgd:d}
	\end{subfigure}
	\hfill\hfill\hfill
	\caption{The loss landscape when optimizing a neural network on the CIFAR-10 data set, perturbed with PGD and different perturbation sizes ($\epsilon$).
		The legend is detailed in the caption of Figure~\ref{fig:mnist_fgsm}.}
	\label{fig:cifar_pgd}
\end{figure*}

We present the results from training a model using both convolution and fully connected layers on the CIFAR-10 data set.
The network consists of three convolutional layers with kernel size 3, a fully connected layer with 32 hidden units and a logits layer of size 10.
All activation functions are ReLU, the loss is cross-entropy and batch normalization is used in the convolutional layers.
The meta-optimizer is trained using a learning rate of $0.001$ determined through grid search.

Since there are striking differences between the convolution and the linear layer, we use two sets of parameters for the meta-optimizer~--~one for optimizing all convolutional layers and one for the linear layers.
Moreover, since the CIFAR-10 problem is more difficult, we train the optimizee for 1000 steps and present the results in Figure~\ref{fig:cifar_fgsm} for perturbations generated with \ac{FGSM} and in Figure~\ref{fig:cifar_pgd} for perturbations generated with \ac{PGD}.
The evaluation was performed as earlier, using 2-fold cross validation for training a meta-optimizer in adversarial settings and transfer it to training a model in normal settings which is tested with adversarial examples.

We observe that in the case of \ac{FGSM}, the transferred meta-optimizer (label L2L-Transfer, Figure~\ref{fig:cifar_fgsm}) exhibits similar behavior as in the MNIST experiments: it has similar and sometimes better performance than ADAM. 
We remind that, in this case, no adversarial examples are used during training.
The meta-optimizer trained normally, but tested with adversarial examples (Transfer-NOT label, Figure~\ref{fig:cifar_fgsm}) performs visibly worse, which strengthens the observation that meta-learning optimization is domain specific.

Figure~\ref{fig:cifar_pgd} shows results from running the same experiment using perturbations generated with the \ac{PGD} method, with a number of 7 steps, as in the original paper~\cite{madry2017towards}.
In all cases, the loss improvements are small, although the meta-optimizer exhibits better performance than ADAM both during training and testing.
However, the improvements in training time are significant since after training an adversarial meta-optimizer, it can be applied to different models without the need to execute 7 forward and back propagation steps for each batch of data.


%
\section{Discussion}
\label{sec:discussion}



Typically used to rapidly adapt to new tasks or generalize outside the i.i.d assumption, meta-learning algorithms show promising results to reduce the training samples needed for adversarial training.
The results presented in this paper suggest these algorithms can be used in the future to build adversarial defenses with less computational resources and capable to adapt to new data.

We hereby note some weaknesses discovered during the process.
Although capable of achieving better performance than hand crafted optimizers~\cite{andrychowicz2016learning,li2017learning} and, as discussed in Section~\ref{sec:experiments}, showing promising results in transferring information about adversarial examples, meta learning algorithms still suffer from broader generalization.
In particular, trained optimizers can not generalize to different activation functions, or between architectures with noticeable differences~\cite{wichrowska2017learned}.
This means that an optimizer trained for ReLU can not be used for sigmoid (or other) activation functions.
Moreover, if the meta-optimizer is not trained with specific data that will be later used, it does not exhibit general behavior. 
For example, if the meta-optimizer does not use any adversarial examples during training, but it encounters such examples during testing, it faces difficulties.
This behavior is illustrated in the figures above with the label Transfer-NOT.

Second order information (taking the gradient of the gradient, as introduced in Section~\ref{sec:formal}) was not used in this paper.
As shown in~\cite{wichrowska2017learned}, this information can help the meta-optimizer better generalize and induce more stable behavior.
However, it also introduces more complexity.
Analyzing the trade-off between the optimizer's complexity and its ability to learn and transfer knowledge related to adversarial vulnerabilities is left for future work.

\section{Conclusions and Future Research}
\label{sec:conclusions}
We introduce a method to learn how to optimize \ac{ML} models robust to adversarial examples, in low data regimes.
Instead of specifying custom regularization terms, they are learned automatically by an adaptive optimizer.
Acquiring meta information about the optimization landscape under adversarial constraints allows the optimizer to reuse it for new tasks.

For future research, we propose to train the meta-optimizer concomitantly with different perturbation types~--~\eg~$l_{1}, l_{2}$-norm~--~and test if the optimizer can learn to robustly optimize under all constraints.
Other perturbations, such as naturally occurring perturbations~\cite{hendrycks2019benchmarking} can also be included.
Another  research direction is to use the meta-optimizer to refine a trained model  and evaluate if it is possible to robustly regularize it with less data.

%
%
%
 \bibliography{bibliography}

\begin{thebibliography}{10}
\providecommand{\url}[1]{\texttt{#1}}
\providecommand{\urlprefix}{URL }
\providecommand{\doi}[1]{https://doi.org/#1}

\bibitem{andrychowicz2016learning}
Andrychowicz, M., Denil, M., Gomez, S., Hoffman, M.W., Pfau, D., Schaul, T.,
  Shillingford, B., De~Freitas, N.: Learning to learn by gradient descent by
  gradient descent. In: NeurIPS. pp. 3981--3989 (2016)

\bibitem{athalye2018obfuscated}
Athalye, A., Carlini, N., Wagner, D.: Obfuscated gradients give a false sense
  of security: Circumventing defenses to adversarial examples. ICML  (2018)

\bibitem{bubeck2018adversarial}
Bubeck, S., Price, E., Razenshteyn, I.: Adversarial examples from computational
  constraints. arXiv:1805.10204  (2018)

\bibitem{oneval}
Carlini, N., Athalye, A., Papernot, N., Brendel, W., Rauber, J., Tsipras, D.,
  Goodfellow, I.J., Madry, A., Kurakin, A.: On evaluating adversarial
  robustness. arXiv:1902.06705  (2019), \url{http://arxiv.org/abs/1902.06705}

\bibitem{carlini2017towards}
Carlini, N., Wagner, D.: Towards evaluating the robustness of neural networks.
  In: 2017 IEEE Symposium on Security and Privacy (S\&P). pp. 39--57. IEEE
  (2017)

\bibitem{cullina2018pac}
Cullina, D., Bhagoji, A.N., Mittal, P.: Pac-learning in the presence of evasion
  adversaries. arXiv:1806.01471  (2018)

\bibitem{finn2017model}
Finn, C., Abbeel, P., Levine, S.: Model-agnostic meta-learning for fast
  adaptation of deep networks. In: ICML. pp. 1126--1135 (2017)

\bibitem{goodfellow2014explaining}
Goodfellow, I.J., Shlens, J., Szegedy, C.: Explaining and harnessing
  adversarial examples. arXiv:1412.6572  (2014)

\bibitem{hendrycks2019benchmarking}
Hendrycks, D., Dietterich, T.: Benchmarking neural network robustness to common
  corruptions and perturbations. ICLR  (2019)

\bibitem{huang2018safety}
Huang, X., Kroening, D., Kwiatkowska, M., Ruan, W., Sun, Y., Thamo, E., Wu, M.,
  Yi, X.: Safety and trustworthiness of deep neural networks: A survey.
  arXiv:1812.08342  (2018)

\bibitem{jacobsen2019exploitint}
Jacobsen, J.H., Behrmann, J., Carlini, N., Florian~Tramer, N.: Exploiting
  excessive invariance caused by norm-bounded adversarial robustness. ICLR
  (2019)

\bibitem{jacobsen2018excessive}
Jacobsen, J.H., Behrmann, J., Zemel, R., Bethge, M.: Excessive invariance
  causes adversarial vulnerability. ICLR  (2019)

\bibitem{katz2017reluplex}
Katz, G., Barrett, C., Dill, D.L., Julian, K., Kochenderfer, M.J.: Reluplex: An
  efficient smt solver for verifying deep neural networks. In: CAV. pp.
  97--117. Springer (2017)

\bibitem{li2016learning}
Li, K., Malik, J.: Learning to optimize. arXiv:1606.01885  (2016)

\bibitem{li2017learning}
Li, K., Malik, J.: Learning to optimize neural nets. arXiv:1703.00441  (2017)

\bibitem{madry2017towards}
Madry, A., Makelov, A., Schmidt, L., Tsipras, D., Vladu, A.: Towards deep
  learning models resistant to adversarial attacks. ICLR  (2018)

\bibitem{mirman2018differentiable}
Mirman, M., Gehr, T., Vechev, M.: Differentiable abstract interpretation for
  provably robust neural networks. ICML  (2018)

\bibitem{miyato2015distributional}
Miyato, T., Maeda, S.i., Koyama, M., Nakae, K., Ishii, S.: Distributional
  smoothing with virtual adversarial training. arXiv:1507.00677  (2015)

\bibitem{ravi2016optimization}
Ravi, S., Larochelle, H.: Optimization as a model for few-shot learning. ICLR
  (2017)

\bibitem{rice2020overfitting}
Rice, L., Wong, E., Kolter, J.Z.: Overfitting in adversarially robust deep
  learning. arXiv preprint arXiv:2002.11569  (2020)

\bibitem{santoro2016meta}
Santoro, A., Bartunov, S., Botvinick, M., Wierstra, D., Lillicrap, T.:
  Meta-learning with memory-augmented neural networks. In: ICML. pp. 1842--1850
  (2016)

\bibitem{schmidhuber1993neural}
Schmidhuber, J.: A neural network that embeds its own meta-levels. In: ICNN.
  pp. 407--412. IEEE (1993)

\bibitem{shafahi2019adversarial}
Shafahi, A., Najibi, M., Ghiasi, A., Xu, Z., Dickerson, J., Studer, C., Davis,
  L.S., Taylor, G., Goldstein, T.: Adversarial training for free!
  arXiv:1904.12843  (2019)

\bibitem{tsipras2018there}
Tsipras, D., Santurkar, S., Engstrom, L., Turner, A., Madry, A.: There is no
  free lunch in adversarial robustness (but there are unexpected benefits).
  arXiv:1805.12152  (2018)

\bibitem{wichrowska2017learned}
Wichrowska, O., Maheswaranathan, N., Hoffman, M.W., Colmenarejo, S.G., Denil,
  M., de~Freitas, N., Sohl-Dickstein, J.: Learned optimizers that scale and
  generalize. In: ICML. pp. 3751--3760 (2017)

\bibitem{wong2017provable}
Wong, E., Kolter, J.Z.: Provable defenses against adversarial examples via the
  convex outer adversarial polytope. ICML  (2018)

\bibitem{wong2020fast}
Wong, E., Rice, L., Kolter, J.Z.: Fast is better than free: Revisiting
  adversarial training. arXiv preprint arXiv:2001.03994  (2020)

\bibitem{younger2001meta}
Younger, A.S., Hochreiter, S., Conwell, P.R.: Meta-learning with
  backpropagation. In: IJCNN. vol.~3. IEEE (2001)

\bibitem{zhang2019you}
Zhang, D., Zhang, T., Lu, Y., Zhu, Z., Dong, B.: You only propagate once:
  Accelerating adversarial training via maximal principle. arXiv:1905.00877
  (2019)

\bibitem{zhang2019limitations}
Zhang, H., Chen, H., Song, Z., Boning, D., Dhillon, I.S., Hsieh, C.J.: The
  limitations of adversarial training and the blind-spot attack. ICLR  (2019)

\end{thebibliography}
 \bibliographystyle{splncs04}

\end{document}